\documentclass[letterpaper, 10 pt, conference]{ieeeconf}  

\IEEEoverridecommandlockouts                              

\usepackage[l2tabu,orthodox]{nag}

\usepackage[
    backend=biber,
    style=ieee,
    sorting=none,
    natbib=true,
    doi=false,
    isbn=false,
    url=false,
    eprint=false,
    maxcitenames=1,
    mincitenames=1
]{biblatex}

\usepackage{acro}
\DeclareAcronym{ICP}{
  short = ICP ,
  long  = iterative closest point,
  tag = acronym
}

\DeclareAcronym{FMCW}{
  short = FMCW ,
  long  = frequency-modulated continuous wave,
  tag = acronym
}

\DeclareAcronym{SOTA}{
  short = SOTA ,
  long  = state-of-the-art,
  list  = State-of-the-Art,
  tag = acronym
}

\DeclareAcronym{TR}{
  short = T\&R ,
  long  = Teach and Repeat,
  tag = acronym
}

\DeclareAcronym{LTR}{
  short = LT\&R ,
  long  = Lidar Teach and Repeat,
  tag = acronym
}

\DeclareAcronym{DA-ICP}{
  short = DA-ICP ,
  long  = Degeneracy-Aware ICP,
  tag = acronym
}

\DeclareAcronym{GPS}{
  short = GPS ,
  long  = Global Positioning System,
  tag = acronym
}

\DeclareAcronym{RANSAC}{
  short = RANSAC ,
  long  = random sample and consensus,
  list  = Random Sample and Consensus,
  tag = acronym
}

\DeclareAcronym{RMSE}{
  short = RMSE ,
  long  = root-mean-squared error,
  list  = Root Mean Squared Error,
  tag = acronym
}

\DeclareAcronym{PCA}{
  short = PCA ,
  long  = principal component analysis,
  list  = Principal Component Analysis,
  tag = acronym
}

\DeclareAcronym{WNOA}{
  short = WNOA ,
  long  = white-noise-on-acceleration,
  list  = White-Noise-on-Acceleration,
  tag = acronym
}

\DeclareAcronym{IMU}{
  short = IMU ,
  long  = Inertial Measurement Unit,
  tag = acronym
}

\DeclareAcronym{UGV}{
  short = UGV ,
  long  = unmanned ground vehicle,
  list  = Unmanned Ground Vehicle,
  tag = acronym
}

\DeclareAcronym{GNSS}{
  short = GNSS ,
  long  = Global Navigation Satellite System,
  tag = acronym
}

\DeclareAcronym{OG}{
  short = OG ,
  long  = odometer-gyroscope,
  list  = Odometer-Gyroscope,
  tag = acronym
}

\DeclareAcronym{BCH}{
  short = BCH,
  long = Baker-Campbell-Hausdorff,
  tag = acronym
}

\DeclareAcronym{CSA}{
  short = CSA,
  long = Canadian Space Agency,
  tag = acronym
}

\DeclareAcronym{UTIAS}{
  short = UTIAS,
  long = University of Toronto Institute of Aerospace Studies,
  tag = acronym
}

\DeclareAcronym{DOF}{
  short = DOF,
  long = Degrees of Freedom,
  tag = acronym
}

\DeclareAcronym{SLAM}{
  short = SLAM,
  long = Simultaneous Localization and Mapping,
  tag = acronym
}

\usepackage{pifont}
\newcommand{\xmark}{\ding{55}}%

\usepackage[pdftex,colorlinks]{hyperref}

\usepackage{siunitx}
\usepackage[all]{nowidow}

\usepackage{tikz}
\usetikzlibrary{arrows}
\usetikzlibrary{arrows.meta}
\usepackage{makecell}

\usepackage{lipsum}

\usepackage{xspace} 

\usepackage{epstopdf}

\usepackage{import}

\usepackage[font=small]{caption}
\usepackage[font=small]{subcaption}

\usepackage{booktabs}

\usepackage{tabularx}
\usepackage{multirow, multicol}

\usepackage{siunitx}

\usepackage{amssymb,amsfonts,amsmath,amscd}

\usepackage{bm}

\newcommand{\bbm}{\begin{bmatrix}}
\newcommand{\ebm}{\end{bmatrix}}

\newcommand{\ignore}[1]{}

\newcommand{\bma}[1]{\left[\begin{array}{#1}}
\newcommand{\ema}{\end{array}\right]}

\DeclareMathAlphabet{\mbf}{OT1}{ptm}{b}{n}

\def\fdotb{{\raisebox{-0.6ex}{ \kern0.2ex\raisebox{0.8ex}{\tiny $\hspace*{-1ex}\circ$}}}}
\def\fddotb{{\raisebox{-0.6ex}{ \kern0.2ex\raisebox{0.8ex}{\tiny $\hspace*{-1ex}\circ\circ$}}}}

\newcommand{\utimes}{ {\raisebox{-0.6ex}{ \kern-1.0ex\raisebox{0.6ex}{ \small $\mathsf{v}$}}} } %
\newcommand{\beq}{\begin{equation}}
\newcommand{\eeq}{\end{equation}}
\newcommand{\bdis}{\begin{displaymath}}
\newcommand{\edis}{\end{displaymath}}
\newcommand{\beqarray}{\begin{eqnarray}}
\newcommand{\eeqarray}{\end{eqnarray}}
\newcommand{\beqarraynn}{\begin{eqnarray*}}
\newcommand{\eeqarraynn}{\end{eqnarray*}}

\DeclareMathAlphabet{\mbf}{OT1}{ptm}{b}{n}

\usepackage{balance}
\usepackage[T1]{fontenc}

\usepackage{array}

\title{\textbf{
   Balancing Act: Trading Off Odometry and Map Registration for Efficient Lidar Localization}
}%

\author{
    Katya M. Papais,
    Daniil Lisus,
    Cedric Le Gentil,
    David J. Yoon, and
    Timothy D. Barfoot
    \thanks{
        \hspace*{-1em}
        This work was supported by the OGS Program provided by the Province of Ontario, and the NSERC CGS D scholarship.
        \newline
        All authors are affiliated with the University of Toronto Institute for Aerospace Studies, 4925 Dufferin Street, Toronto, Ontario, Canada. Corresponding author: \texttt{katya.papais@robotics.utias.utoronto.ca}
    }%
}%

\begin{document}

\maketitle
\thispagestyle{empty}
\pagestyle{empty}

\begin{abstract}
Most autonomous vehicles rely on accurate and efficient localization, which is achieved by comparing live sensor data to a preexisting map, to navigate their environment. 
Balancing the accuracy of localization with computational efficiency remains a significant challenge, as high-accuracy methods often come with higher computational costs. 
In this paper, we present two ways of improving lidar localization efficiency and study their impact on performance. 
First, we integrate two lightweight odometry estimators, a correspondence-free Doppler-inertial estimator and a low-cost wheel odometer-gyroscope (OG) method, into a topometric localization pipeline and compare them against a state-of-the-art (SOTA) iterative closest point (ICP) baseline. 
We highlight the trade-offs between these approaches: the Doppler and OG estimators offer faster, lightweight updates, while ICP provides higher accuracy at the cost of increased computational load. 
Second, by controlling the frequency of localization updates and leveraging odometry estimates between them, we demonstrate that accurate localization can be maintained while optimizing for computational efficiency using any of the presented methods. 
We evaluate these approaches using over $100 \ \unit{\km}$ of unique real-world driving data in different on-road environments.
By varying the localization interval, we demonstrate that computational effort can be reduced by $27\%$, $80\%$, and $91\%$ for the ICP, Doppler, and OG estimators, respectively, while maintaining SOTA accuracy.
\end{abstract}





\section{Introduction}
\label{sec:intro}

A critical part of autonomous vehicle (AV) navigation is the ability to localize the vehicle in a known map of the environment. 
This task, called localization, is crucial for down-stream AV objectives such as trajectory planning, obstacle tracking and avoidance, and smooth vehicle handling. 
\begin{figure}[ht!]
    \centering
    \includegraphics[width=1.0\linewidth]{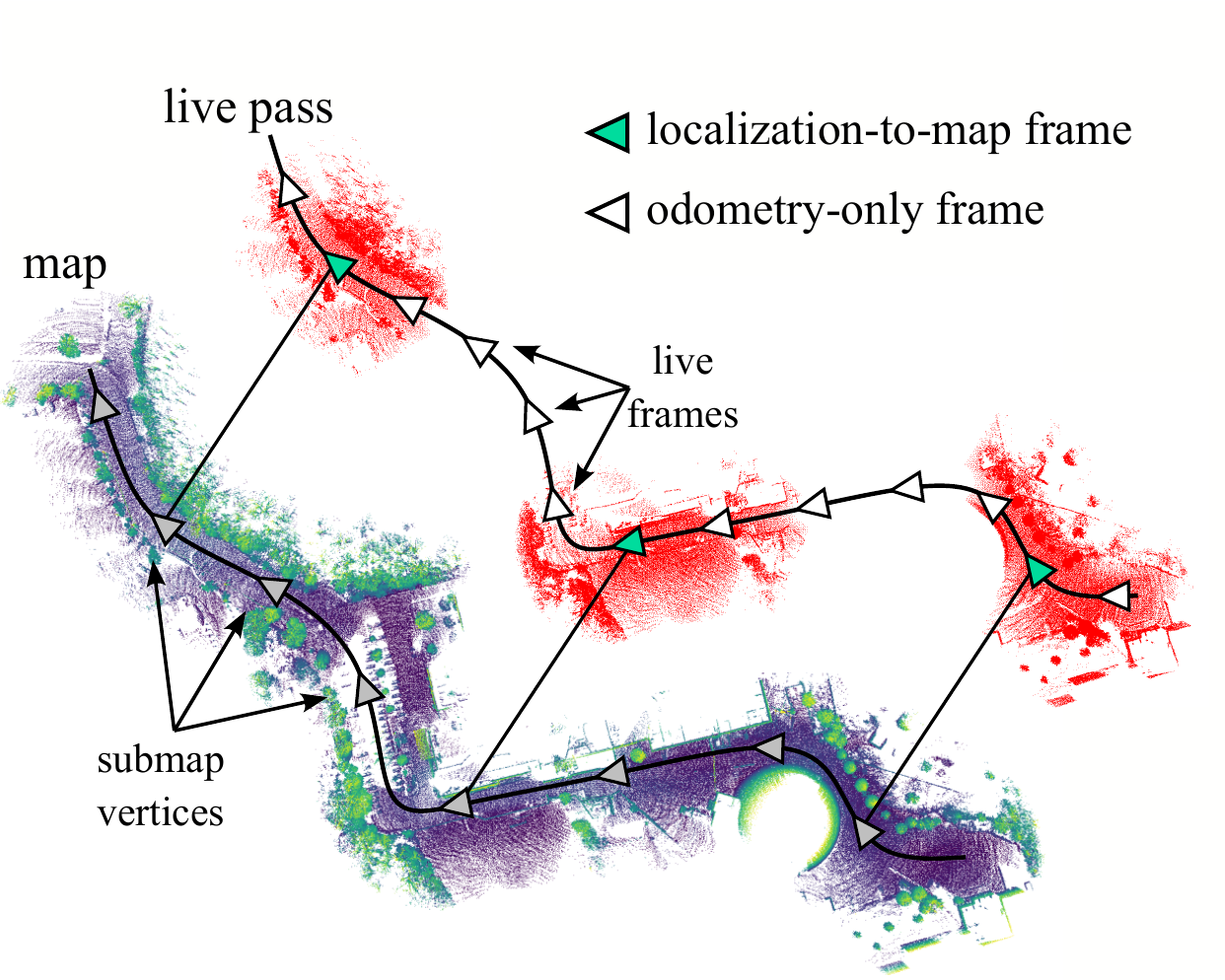}
    \caption{We explore the trade-off between computational efficiency and localization accuracy with infrequent map matching, analyzing how dead-reckoning between these localization attempts impacts overall performance.}
    \label{fig:eye_candy}
\end{figure}
A map-based localization pipeline will conventionally contain two parts: a frame-to-frame, relative-motion odometry estimator, and a frame-to-map localization estimator \citep{Burnett2022Ready, KumarSurvey}. 
A common technique employed in both estimators when using a lidar sensor is the \ac{ICP} algorithm, which iteratively aligns two point clouds \citep{Besl1992AMF, Pomerleau2015ARO}. 
The odometry estimator matches live lidar point clouds between consecutive frames, while the localization estimator matches live lidar point cloud to a point cloud map, typically constructed using lidar data from a previous traversal. 
To get \ac{SOTA} localization performance, dense point clouds and many ICP iterations are required, creating an inherent trade-off between accuracy and computational efficiency.
 
Emerging \ac{FMCW} lidar sensors offer new opportunities for efficient and geometry-independent motion estimation. 
We leverage FMCW lidar to implement a correspondence-free odometry estimator, which uses relative radial velocity and gyroscope measurements to estimate vehicle motion, similar to \citep{Wu2022PickingUS, YoonN4S}. 
Notably, this Doppler-inertial odometry method, referred to throughout as Doppler odometry, can remain robust in feature-sparse environments, such as highways, tunnels, and open fields, where as traditional ICP-based methods may struggle due to the lack of geometric structure. 
We additionally utilize a lightweight wheel \ac{OG} method that removes the expense of processing lidar data altogether \citep{LeGentil2025}. 
Since these two estimators are not dependent on scene geometry, they are robust in environments where geometry-based ICP struggles and offer a computational advantage due to lower processing overhead and the elimination of point cloud association.

This paper presents the study of computational efficiency in lidar-based navigation by evaluating Doppler, OG, and ICP odometry within a unified localization pipeline. 
We investigate the trade-off space between runtime and accuracy by systematically varying the interval between map-matching updates, relying on odometry for dead-reckoning in the interim.
This approach, illustrated in Figure~\ref{fig:eye_candy}, is validated through experiments spanning from computationally inexpensive, infrequent map matching enabled by OG or Doppler odometry to high-frequency, resource intensive ICP updates at every frame.
Our contributions are summarized as follows:
\begin{itemize}
    \item We present the integration of Doppler and OG odometry methods into a modular lidar localization pipeline as efficient alternatives to an ICP-based method.
    \item We study the trade-off between computational effort and localization performance by varying the frequency of localization updates in three odometry algorithms using over $100 \ \unit{km}$ of unique real-world driving data.
\end{itemize}

The rest of this paper is structured as follows. Section \ref{sec:RW} presents relevant prior work. Section \ref{sec:methodology} discusses the details of the implemented pipelines. Section \ref{sec:experiments} shows and discusses the experimental results. Section \ref{sec:conclusion} highlights the main conclusions.

\section{Related Work}
\label{sec:RW}

\subsection{ICP-based Lidar Odometry}
ICP-based lidar odometry estimates the relative transformation between two point clouds by iteratively associating points using nearest-neighbour search \citep{Besl1992AMF, Pomerleau2015ARO}. 
Since ICP-based methods can become computationally expensive when using dense lidar point clouds, many SOTA methods implement optimizations, such as voxel downsampling, to achieve real-time performance \citep{VizzoKISS, DellenbachCTICP, ChenLIOCT}. 
%
A limitation to conventional ICP-based methods is the discrete-frame assumption\footnote{%
    A frame is defined as the set of points collected over a single, full lidar scan.%
}, which can introduce misalignment when motion occurs throughout a frame.
To address this, continuous-time ICP methods estimate the trajectory as a smooth function of time, allowing pose estimates to vary throughout each frame \citep{DellenbachCTICP, ChenLIOCT, Wu2022PickingUS}. 
The ICP-based odometry method used in this work is most similar to the CT-ICP method in \citep{DellenbachCTICP}, leveraging a continuous-time representation to improve robustness. 

Another challenge for conventional ICP methods is geometrically degenerate environments, where a lack of distinct features can lead to estimation failure.
This is typically overcome through the integration of additional sensors, such as inertial measurement units (IMU) \citep{YeTCLIO, ChenLIOCT, Zhao2021SuperOI, Shan2020LIOSAMTL, ZhangLOAM, BurnettCTLIO}.
In this work, we include gyroscope measurements directly in our ICP-based odometry optimization problem, and undistort point clouds using the posterior of our estimated trajectory, similar to \cite{BurnettCTLIO}.


\subsection{ICP-based Lidar Localization}
Global Navigation Satellite System (GNSS) sensors are commonly used in combination with lidar sensors to perform vehicle localization. 
For example, \citet{Yoneda2014} use ICP to align live lidar point clouds to a geo-referenced map to study the use of mobile mapping systems in AV navigation. 
However, GNSS positional accuracy degrades in complex or rapidly changing environments, making it unreliable in urban canyons or tunnels. 
In such cases, topometric localization offers a solution by matching onboard sensor data to a prebuilt map of the environment \citep{Badino2011, Barfoot2010, Burnett2022Ready, Landry2016}. This paper follows the work of \citet{Burnett2022Ready}, performing ICP-based localization within a pose-graph-based topometric map. This approach allows to incorporate measurements as they arrive. 

\subsection{Lidar Navigation Efficiency}
Several papers discuss and attempt to improve the trade-off between computational effort and performance in lidar-based localization and other lidar-based navigation tasks \citep{niedzwiedzki2020real, ZhangLOAM, LEGOLOAM}. \citet{niedzwiedzki2020real} present a real-time lidar localization algorithm for GPS-denied environments. The algorithm applies an extended Kalman filter to each lidar point and localizes a robot relative to a triangular mesh reference map. This approach creates a compact map representation, thus reducing computational cost, without a significant reduction in accuracy. The work by \citet{ZhangLOAM} is most similar to ours in that fast, low-fidelity odometry estimates at $10 \ \unit{\hertz}$ are refined by slower, high-fidelity localization updates at $1 \ \unit{\hertz}$. While this approach is common in simultaneous localization and mapping (SLAM) systems, where continuous map-building and updating are essential to achieve real-time \citep{ZhangLOAM, LEGOLOAM}, our work differs in that we use a prebuilt reference map and analyze the trade-offs associated with achieving SOTA localization; however, the idea of restricting the frequency of localization updates to optimize the associated trade-offs can be applied to other localization pipelines, such as \citep{BClaraco2025}. To the best of our knowledge, none have showed the impact of infrequent map matches or discussed how to select an optimal frequency for optimizing computational efficiency and accuracy in a localization pipeline. 

\subsection{FMCW Sensors}
The use of FMCW technology and Doppler-based motion estimation is well-established in the radar domain \citep{Vivet2013LocalizationAM, Kellner2013, Kramer2020, Park2021, Kubelka2024, GaoDCLOC, lisus2024doppler}; its application to lidar is more recent \citep{Hexsel2022DICPDI, Wu2022PickingUS, YoonN4S, Zhao2024, PangFMCW}.
FMCW lidar sensors provide relative per-point radial velocity data via the Doppler effect in addition to 3D spatial information, and offer enhanced range resolution and resistance to interference compared to traditional time-of-flight lidar \citep{Behroozpour2017LidarSA}. 

Recent work has demonstrated that incorporating these velocity measurements into ICP-based methods improves estimation robustness  in geometrically degenerate environments \citep{Hexsel2022DICPDI, Wu2022PickingUS}.
\citet{Hexsel2022DICPDI} first showed that Doppler velocity measurements can be used for motion estimation methods using the Doppler-ICP algorithm. 
\citet{Wu2022PickingUS} built upon this by incorporating a continuous-time estimator and eliminating the need for external motion compensation when preprocessing data. 

\citet{YoonN4S} showed that data association between point clouds can be eliminated entirely with their lightweight, correspondence-free odometry solution. 
The quantitative results show significant computational savings compared to SOTA ICP-based odometry while maintaining reasonable accuracy. 
In this work, we integrate this correspondence-free algorithm into a topometric localization pipeline, allowing for a comparative analysis of its performance against ICP and proprioceptive baselines across a range of map-matching intervals.

\section{Methodology}
\label{sec:methodology}

\subsection{Problem Formulation}
\begin{figure}[t]
    \centering
    \begin{subfigure}[t]{\linewidth}
        \centering
        \includegraphics[width=\textwidth]{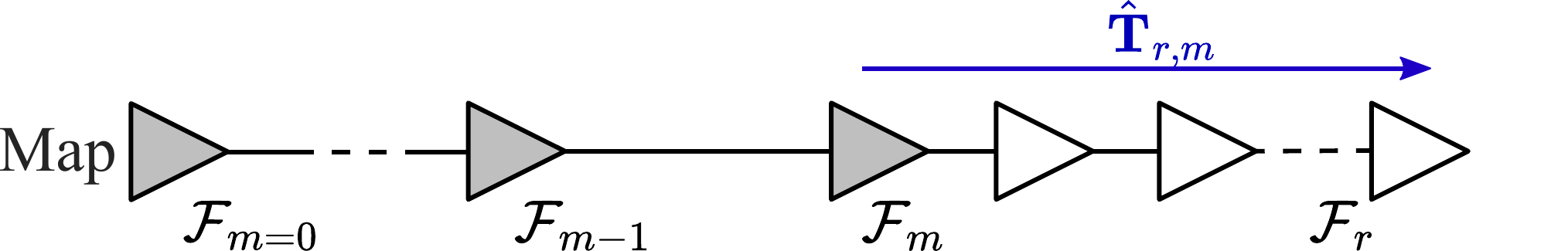}
        \caption{Submap Construction}
        \label{sfig:submap_constr}
    \end{subfigure}
    \hfill
    \begin{subfigure}[t]{\linewidth}
        \centering
        \includegraphics[width=\textwidth]{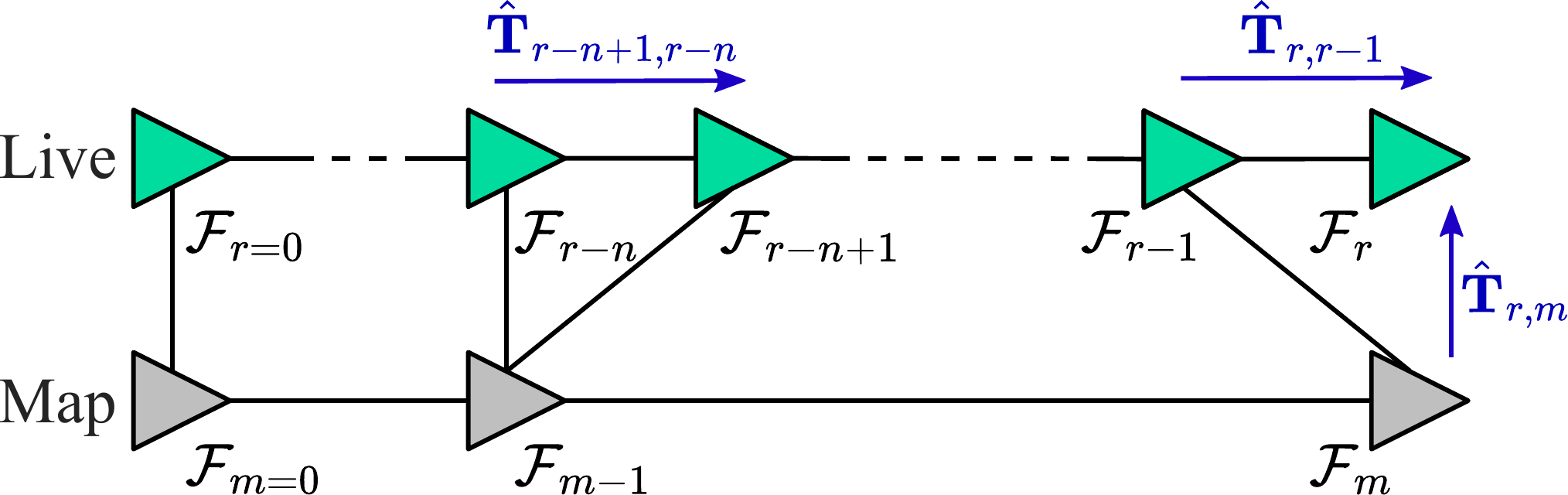}
        \caption{Localization}
        \label{sfig:loc}
    \end{subfigure}
    \hfill
    \begin{subfigure}[t]{\linewidth}
        \centering
        \includegraphics[width=\textwidth]{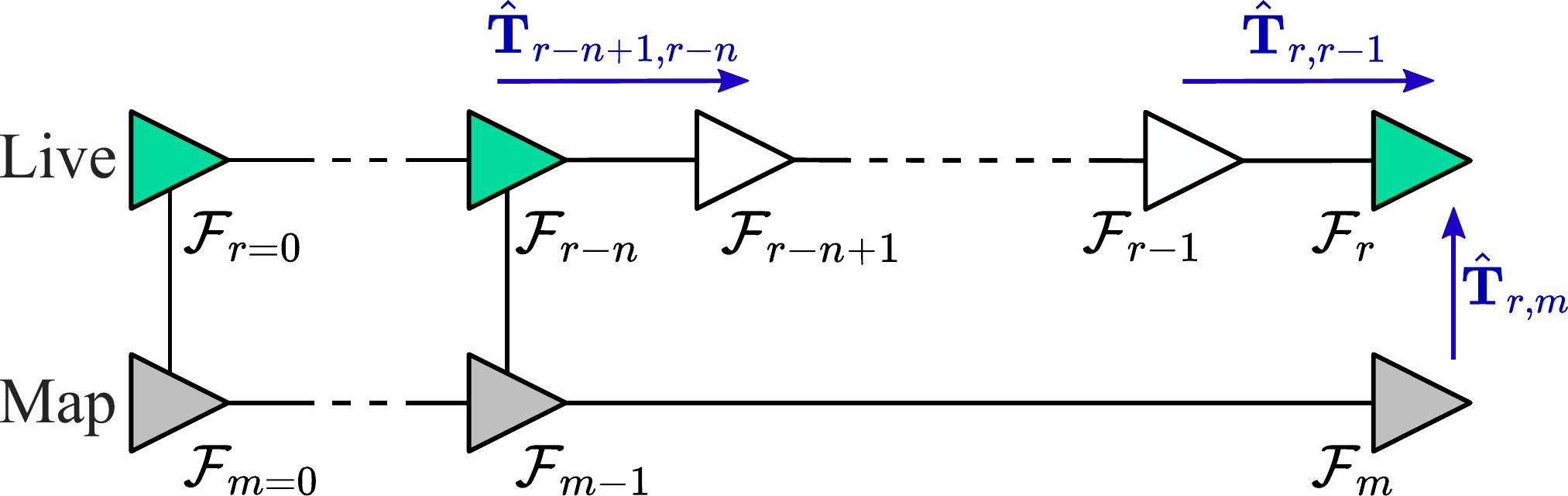}
        \caption{Infrequent Localization}
        \label{sfig:infreqloc}
    \end{subfigure}
        \caption{The structure of the pose graph during submap construction and localization, illustrating the localization pipeline from \cite{Burnett2022Ready} and the modified pipeline used in this work. White vertices represent odometry-only frames, while green vertices represent localization-to-map frames.
        }
    \label{fig:graph}
\end{figure}
We use the teach and repeat framework as our localization pipeline \citep{Barfoot2010, Paton2016, Paton2017ICS, Burnett2022Ready}. During the initial mapping pass, the vehicle collects sensor data along a route, which is used to construct local submaps stored as vertices in a pose graph. These vertices are connected by edges representing relative pose estimates based on odometry. Fig.~\ref{fig:eye_candy} and \ref{fig:graph} show odometry estimates and submap vertices as white and grey nodes, respectively. 

During the localization pass, the vehicle follows the same route, again using odometry estimates to build a new branch of the pose graph while simultaneously localizing against the previously recorded submaps. This localization step corrects for odometric drift without relying on global positioning. Odometry estimates that are localized against a submap are coloured green in Fig.~\ref{fig:eye_candy} and \ref{fig:graph}.

We use the following notation:
\begin{itemize}
    \item $r$ denotes the current frame being estimated,
    \item $m$ denotes the submap vertex from the teach pass, 
    \item $n$ denotes the number of frames between successive attempts to localize,
    \item $\hat{\mathbf{T}}_{r,r-1}$ denotes the latest odometry estimate from the previous frame $\boldsymbol{\mathcal{F}}_{r-1}$, to the current frame $\boldsymbol{\mathcal{F}}_{r}$,
    \item $\hat{\mathbf{T}}_{r,m}$ denotes the localization estimate from the submap vertex $\boldsymbol{\mathcal{F}}_{m}$, to the current frame, $\boldsymbol{\mathcal{F}}_{r}$.
\end{itemize}

The goal of odometry is to estimate $\hat{\mathbf{T}}_{r,r-1}$, the relative transformation between sequential frames. 
The goal of localization is to estimate $\check{\mathbf{T}}_{r,m}$, the global transformation from the reference map to the current frame.

\citet{Burnett2022Ready} attempt localization at every frame ($n = 1$) using only ICP for odometry (see Fig.~\ref{sfig:loc}).
We integrate two new lightweight odometry modules, Doppler and OG odometry, to study their specific impacts in a localization pipeline.
By progressively increasing $n$ while using these different estimators, we evaluate the resulting effect on computational efficiency and localization accuracy (see Fig.~\ref{sfig:infreqloc}).
The following sections describe the three odometry methods and the localization procedure considered in this paper.

\subsection{Doppler Odometry}
\label{sec:dop_odometry}
We use the Doppler odometry methodology as presented by \citet{YoonN4S}. 
We provide a summary here and refer the reader to \cite{YoonN4S} for further details. 

\subsubsection{Raw Data Preprocessing}
Point measurements are downsampled uniformly in azimuth-elevation space and corrected for range-dependent bias using an offline-trained linear regression model. 
Doppler measurement outliers, which can be caused by moving objects or reflections in the scene, are rejected using random sample and consensus (RANSAC) \cite{FisRANSAC}. 
In this work, we supplement this with an additional outlier rejection threshold of $3 \ \text{m}/\text{s}$ on the change in forward velocity between frames. 
This ensures temporal consistency and prevents RANSAC from incorrectly latching onto dynamic objects when static points are sparse, as done in \citep{KellnerDopRadar, lisus2024doppler} for Doppler-based estimation with radar.

\subsubsection{Estimation}
\label{ssec:doppler_est}
This method formulates the velocity estimation as an exactly sparse Gaussian process (GP) regression problem, based on the work by \citet{barfoot2014batch}.
The estimator solves for the continuous-time 6-DOF vehicle velocity, $\boldsymbol{\varpi}(t)$, by minimizing a cost function that incorporates a \ac{WNOA} motion prior, vehicle kinematics, gyroscope measurements, and per-point Doppler measurements.
Since these factors are formulated as linear models with respect to the velocity state, the objective function is a linear least-squares problem. 
This allows the state to be recovered using a single, computationally efficient linear solve.
The relative pose is then approximated by numerically integrating the estimated velocity $\boldsymbol{\varpi}(t)$ over the lidar frame, sampling at small time intervals to construct a sequence of $\textit{SE}(3)$ transformations that represent the motion over the frame. 

\subsection{OG Odometry}
\label{sec:OG_odometry}
As a non-lidar implementation, we adapt the OG odometry method presented by \citet{LeGentil2025} for use in $SE(3)$.
We provide a summary here and refer the reader to \citep{LeGentil2025} for further details. 
Unlike the other odometry methods used in this work, this method does not utilize lidar point cloud data and instead relies exclusively on wheel odometer and gyroscope measurements. 
Thus, there is no lidar data preprocessing required to produce an odometry estimate. 
However, to ensure a controlled comparison when localizing, all pipelines (including OG) preprocess the lidar point clouds for scan-to-map matching using the method presented in Section \ref{subsec:icp_preprocessing}.
For consistency within the localization system, we output odometry estimates at the same timestamps as the lidar frames. 

\subsubsection{Estimation}
The OG method estimates the vehicle's pose by combining distance information from a single rear-wheel encoder with angular velocity measurements from a gyroscope.
The distance travelled increment is computed by converting wheel encoder ticks into meters based on the wheel's circumference.
Simultaneously, it integrates the angular velocities to obtain the rotation increment. 
By applying the distance increments along the current heading, the estimator continuously updates the vehicle's pose relative to its starting point.

\subsection{ICP Odometry Baseline}
\label{sec:icp_odometry}
As a baseline for evaluating the performance of the lightweight odometry methods, we adopt the ICP odometry presented by \citet{Burnett2022Ready}. 
This estimation approach is a widely recognized method for point cloud registration and motion estimation. 
We provide a brief summary here and refer the reader to \cite{Burnett2022Ready} for further detail.

\subsubsection{Raw Data Preprocessing}
\label{subsec:icp_preprocessing}
Each lidar scan is first downsampled using a voxel filter.
Plane features using principal component analysis (PCA); points are retained if their feature score is greater than $0.95$, with the corresponding smallest eigenvector serving as the surface normal for the point-to-plane metric.

\subsubsection{Estimation}
The continuous-time ICP approach extends traditional ICP by representing the trajectory as an exactly sparse GP \cite{Anderson2015}. 
The state $\mathbf{x}(t) = \{ \mathbf{T}(t), \boldsymbol{\varpi}(t)\}$ is estimated by minimizing a non-linear cost function that incorporates the same WNOA motion prior and gyroscope factors as the Doppler estimator from Section \ref{ssec:doppler_est}, in addition to point-to-plane distance residuals. 
This method minimizes the point-to-map residuals weighted by an information matrix derived from surface normals. 
The system is solved iteratively using Gauss-Newton optimization, with nearest-neighbour data associations updated at each iteration until convergence.

\subsection{Mapping}
Reference maps are constructed offline using groundtruth data, following the assumption that high-quality lidar maps already exist. 
While the odometry estimates from each method are capable of producing high-quality maps independently, we utilize groundtruth for map generation to ensure localization results are not biased by the specific odometry method used during the mapping phase.
This standardization allows for a controlled comparison of the three localization pipelines, as each operates against an identical, high-fidelity reference map.

Mapping is achieved by incrementally building a series of local submaps that represent the vehicle's environment.
Using the groundtruth trajectory, the system monitors the vehicle's progress to define the transformation between the latest submap $\boldsymbol{\mathcal{F}}_{m}$, and the current live lidar scan $\boldsymbol{\mathcal{F}}_{r}$ (Fig.~\ref{sfig:submap_constr}).
When the relative motion exceeds a predefined translation or rotation threshold, a new submap vertex $\boldsymbol{\mathcal{F}}_{m+1}$ is added. 
The submap at each new vertex is built by accumulating point clouds from the last three frames.

\subsection{Localization}
To ensure consistency during localization, all pipelines preprocess the point cloud using the same method, as presented in Section \ref{subsec:icp_preprocessing}. 
The point cloud is then undistorted using the odometry estimate from Doppler, OG, or ICP. 
This ensures that the same filtered point cloud is used for localization, with the only difference being the odometry prior. 

Further computational savings are achieved with the Doppler and OG estimators, which do not rely on previously stored point clouds for odometry. 
Instead, the preprocessed point cloud is only stored if a localization attempt is flagged for that frame. 
This significantly reduces memory usage, particularly when localization is performed infrequently. 
The ICP pipeline must retain recent point clouds for use in each scan alignment, which incurs additional storage and computational overhead regardless of localization frequency. 

We use the localization methodology from \cite{Burnett2022Ready}, summarized here for completeness. 
During the localization, ICP is used to align the current motion-compensated lidar frame, $\boldsymbol{\mathcal{F}}_{r}$, to the submap associated with the nearest vertex frame, $\boldsymbol{\mathcal{F}}_{m}$, from the teach pass sequence. 
The nearest vertex is determined using the latest odometry estimate by traversing the edges within the pose graph to find the spatially closest candidate. 
The prior, $\check{\mathbf{T}}_{r,m}$, is constructed using a previous map match and is used to form a prior error term
\begin{equation}
    \mathbf{e}_{\text{prior}} = \ln(\check{\mathbf{T}}_{r,m}^{} \mathbf{T}_{r,m}^{-1})^{\vee},
\end{equation}
where $\mathbf{T}_{r,m}$ is the transform we seek to estimate, and $(\cdot)^{\vee}$ is the $SE(3)$ `vee' operator \cite{Barfoot}. 
The measurement error is
\begin{equation}
    \mathbf{e}_{\text{loc},i} = \mathbf{D}(\mathbf{p}_{i} - \mathbf{T}_{r,m}^{-1} \mathbf{T}_{r,s}^{} \mathbf{q}_i),
\end{equation}
where $\mathbf{q}_i$ is a homogeneous point in the sensor frame from the current motion-compensated point cloud, $\mathbf{p}_{i}$ is a homogeneous point from the $m^\text{th}$ submap associated to $\mathbf{q}_i$, $\mathbf{T}_{r,s}$ is the extrinsic transformation between the sensor and vehicle frames, and $\mathbf{D}$ is a constant selection matrix that removes the fourth element. 

The non-linear cost function we seek to optimize is
\begin{equation} 
\label{eq:defualt_icp}
    \begin{aligned}
    J_{\text{loc}} &= \frac{1}{2} {\mathbf{e}_{\text{prior}}^{}}^{\top} \mathbf{Q}_{\text{prior}}^{-1}  \mathbf{e}_{\text{prior}}^{}
    + \sum_i \frac{1}{2}  \mathbf{e}_{\text{loc},i}^{\top} \mathbf{R}_{i}^{-1}  \mathbf{e}_{\text{loc},i},
    \end{aligned}
\end{equation}
where $\mathbf{Q}_{\text{prior}}$ and $\mathbf{R}_{i}$ are the covariance matrices associated with the prior and measurement error terms, respectively.

\subsubsection{Localize every frame, $n=1$}
When localizing every frame (see Fig.~\ref{sfig:loc}), the initial pose estimate alignment, $\check{\mathbf{T}}_{rm}$, is computed by compounding the transformations between frames in the pose graph:
\begin{equation}
    \check{\mathbf{T}}_{r,m} = \mathbf{T}_{r,r-1}\mathbf{T}_{r-1,m-1}\mathbf{T}_{m-1,m}.
\end{equation}

\subsubsection{Localize every $n$ frames}
Instead of localizing after each frame, we restrict localization to be triggered after a set number of frames, denoted by $n$ (see Fig.~\ref{sfig:infreqloc}). 
Between localization calls, we rely on the odometry estimates and dead-reckon for $n$ frames. 
Once localization is triggered, our prior is computed by compounding the odometry estimates since the last successful localization:
\begin{equation}
    \begin{split}
        \check{\mathbf{T}}_{r,m} = \mathbf{T}_{r,r-1}\mathbf{T}_{r-1,r-2}
         \cdots \mathbf{T}_{r-n,m-1}\mathbf{T}_{m-1,m}.
    \end{split}
\end{equation}
This process reduces computational overhead in the AV navigational pipeline by reducing the number of costly ICP-based map alignment calls.

\section{Experiments}
\label{sec:experiments}
    
\subsection{Data}
We utilize the Boreas Road Trip dataset \cite{Lisus2026BORRT}, which includes data collected by an Aeva Aeries II FMCW lidar sensor. 
This sensor has a $120^{\circ}$ horizontal field of view, a $30^{\circ}$ vertical field of view, a maximum operating range of $500 \ \unit{\m}$, and operates at $10 \ \unit{\hertz}$. 
We also utilize data from the stand-alone Silicon Sensing DMU41 IMU, which operates at $200 \ \unit{\hertz}$, and the Dynapar Wheel Encoder mounted to the rear-left wheel of the vehicle.

We evaluated our approach on five routes totalling approximately $116 \ \unit{\km}$ of unique driving data, representing over $2700 \ \unit{\km}$ of cumulative data processed across all tested configurations and pipelines. 
The \texttt{Suburban} route consists of residential areas with many static features.
The \texttt{Industrial} route covers urban areas with slower speeds and distinct geometric landmarks.
Several sequences in this set were collected during active snowfall.
The \texttt{Regional North} and \texttt{Regional South} routes have high-speed highway segments with fewer static features; these two routes follow the same highway in opposite directions.
Finally, the \texttt{Tunnel} route includes a $840$-$\unit{m}$-long tunnel segment. 

For each route, one holdout sequence was reserved for tuning covariances and biases, one sequence was used to construct the reference map, and three sequences were used for experimental testing. 
Since the \texttt{Regional} routes contain only 3 sequences each, one was used for the reference map and the remaining sequences for localization testing; the reported \texttt{Regional} results are the average of these four trials.  
Tuning was performed across all holdout sequences collectively to prevent overfitting to a specific route. 

\subsection{Results}
\begin{figure*}[t]
    \centering
    \includegraphics[width=\textwidth]{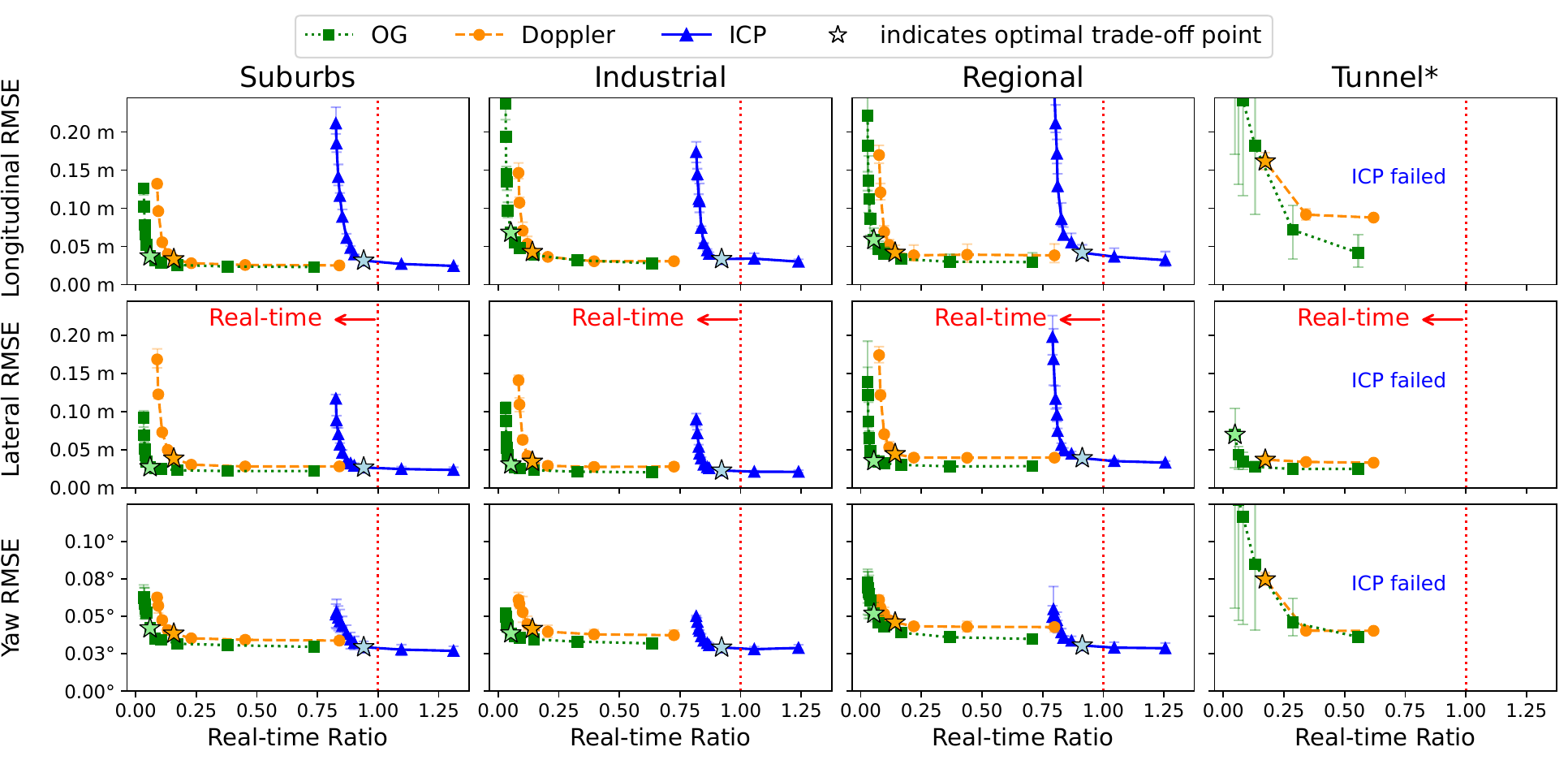}
    \caption{Pareto-front plots for the longitudinal (top), lateral (middle), and yaw (bottom) RMSE, comparing OG, Doppler, and ICP-based pipelines across four routes. The plots show the trade-off between localization error and computational effort, with markers representing each tested interval. The knee point of the curve, marked with a star, is the optimal trade-off point between these two objectives. Error bars show the spread of data at each interval. For the Doppler pipeline, the tested intervals are $n = 1,2,5,10,15,25,50,75$, while the OG and ICP pipelines extend this range to include $n=100,150,200$. ${}^*$Doppler result is averaged over the two successful trials.}
    \label{fig:pareto}
\end{figure*}
\begin{table*}[t]
    \renewcommand{\arraystretch}{1.3}
    \small
    \centering
    \caption{Localization performance summary across test routes. For each route, results are reported for the three estimators at their respective knee points, followed by the best-performing overall configuration (minimum RMSE). Reported position and orientation RMSE, as well as per-frame runtime, are averaged across successful trials.}
    \label{tab:results_summary}
    \begin{tabularx}{\linewidth}{c c c | *{6}{>{\centering\arraybackslash}X} | c}
        \toprule
        \textbf{Route} &
        \makecell{\textbf{Method /} \\ \textbf{Interval ($n$)}} &
        \makecell{\textbf{Runtime} \\ \textbf{[ms]}} &
        \textbf{lat. [m]} &
        \textbf{long. [m]} &
        \textbf{vert. [m]} &
        \textbf{roll [$^\circ$]} &
        \textbf{pitch [$^\circ$]} &
        \textbf{yaw [$^\circ$]} &
        \makecell{\textbf{Successes}} \\
        \hline

        \multirow{4}{*}{\texttt{Suburban}} 
          & OG/25 & 6 & 0.027 & 0.037 & 0.082 & 0.045 & 0.037 & 0.042 & 3/3 \\
          & Doppler/10 & 16 & 0.039 & 0.033 & 0.040 & 0.054 & 0.044 & 0.039 & 3/3 \\
          & ICP/5 & 95 & 0.027 & 0.031 & 0.031 & 0.035 & 0.027 & 0.029 & 3/3 \\
          \cline{2-10}
          & \textbf{Best: ICP/1} & 131 & 0.024 & 0.025 & 0.029 & 0.026 & 0.023 & 0.027 & 3/3 \\
        \hline

        \multirow{4}{*}{\texttt{Industrial}} 
          & OG/25 & 5 & 0.031 & 0.068 & 0.049 & 0.043 & 0.039 & 0.039 & 3/3 \\
          & Doppler/10 & 14 & 0.034 & 0.043 & 0.042 & 0.058 & 0.043 & 0.042 & 3/3 \\
          & ICP/5 & 95 & 0.023 & 0.034 & 0.047 & 0.034 & 0.030 & 0.029 & 3/3 \\
          \cline{2-10}
          & \textbf{Best: ICP/1} & 128 & 0.021 & 0.030 & 0.044 & 0.022 & 0.020 & 0.029 & 3/3 \\
        \hline

        \multirow{4}{*}{\texttt{Regional}} 
          & OG/25 & 5 & 0.035 & 0.059 & 0.094 & 0.047 & 0.038 & 0.052 & 4/4 \\
          & Doppler/10 & 14 & 0.044 & 0.042 & 0.040 & 0.051 & 0.044 & 0.046 & 4/4 \\
          & ICP/5 & 92 & 0.039 & 0.042 & 0.042 & 0.033 & 0.027 & 0.031 & 4/4 \\
          \cline{2-10}
          & \textbf{Best: ICP/1} & 126 & 0.033 & 0.032 & 0.038 & 0.028 & 0.023 & 0.029 & 4/4 \\
        \hline

        \multirow{4}{*}{\texttt{Tunnel}} 
          & OG/25 & 5 & 0.070 & 0.325 & 0.164 & 0.064 & 0.055 & 0.143 & 3/3 \\
          & Doppler/5 & 17 & 0.037 & 0.162 & 0.078 & 0.046 & 0.035 & 0.075 & 2/3 \\
          & ICP/\small \xmark & \small \xmark & \small \xmark & \small \xmark & \small \xmark & \small \xmark & \small \xmark & \small \xmark & 0/3 \\
          \cline{2-10}
          & \textbf{Best: OG/1} & 56 & 0.025 & 0.042 & 0.048 & 0.065 & 0.041 & 0.036 & 3/3 \\
        \bottomrule
    \end{tabularx}%
\end{table*}
\subsubsection{Evaluation Metrics}
Localization performance is quantified using the \ac{RMSE} between predicted relative poses and groundtruth relative poses. 
For each repeated route, we align every localization estimate with the closest associated groundtruth timestamp and compute the $SE(3)$ error. 
Translational errors are reported in meters, and rotational errors are reported in degrees.

Runtime is measured independently and reported as the real-time ratio, defined as the ratio between runtime of the localization pipeline and the collection time of the sequence. 
A compute ratio of $1$ indicates real-time performance; per-frame runtimes below $100 \ \unit{\ms}$ are considered real-time using a $10 \ \unit{\hertz}$ lidar.
All timing results were obtained using a Lenovo ThinkPad P16 Gen 2 (Intel i7-13850HX CPU, 64 GB RAM). 
The system utilizes a multi-threaded CPU implementation (10 threads) and does not leverage GPU acceleration.

We use Pareto curves as a graphical representation of the trade-off between two competing objectives; in our case, localization accuracy and computational effort. 
Fig.~\ref{fig:pareto} shows the performance of our pipelines across varying configurations, showing how accuracy and computational cost change relative to each other as the localization interval increases. 
The `knee point' on the curve corresponds to the most cost-effective trade-off point between computation time and accuracy. 
This knee point is identified as the configuration that minimizes the area of the rectangle formed between the origin and each point on the curve, using the $L_2$ norm of the longitudinal and lateral errors as the accuracy cost and the real-time ratio as the compute cost.
The knee point provides a natural way to select the `optimal' localization frequency.

Table \ref{tab:results_summary} presents results for the \ac{OG}, Doppler, and ICP pipelines at two key configurations: the most accurate across all estimators, and the knee point of each estimator. 
For each configuration, we report the translational and rotational RMSE values, and the average runtime per frame. 

\subsubsection{Estimator Analysis}
The ICP-based estimator reaches its knee point at $n=5$. 
Since the lidar operates at $10 \ \unit{\hertz}$, this corresponds to localizing twice per second. 
At a driving speed of $60 \ \unit{\km / h}$, $n=5$ corresponds to approximately $8.33 \ \unit{\m}$ travelled between localization attempts. 
Although the estimator maintains reasonable accuracy beyond the knee point, reducing the localization frequency beyond it yields diminishing computational savings as the pipeline's runtime becomes bottlenecked by the odometry. 
ICP's robustness against drift allows it to maintain accuracy with fewer corrections, extending its useful range to higher intervals than the Doppler estimator and performing comparably to the OG estimator. 
However, this method struggles to achieve real-time performance when using smaller intervals. 
Using our approach, the ICP-based method achieves real-time performance when $n\geq5$ while maintaining accuracy.

The Doppler estimator reaches its knee point at $n=10$ where the runtime asymptotically approaches the odometry-only baseline of approximately $10 \ \unit{\ms}$ per frame. 
However, because this method relies on integrating radial velocity measurements within a limited FOV, it is more susceptible to cumulative drift. 
While we apply corrections for radial velocity biases, these compensations are imperfect; the errors compound over time, causing accuracy to degrade significantly as the dead-reckoning interval increases until localization becomes unsuccessful beyond $n=75$.

The OG estimator reaches its knee point at $n=25$, which is significantly larger than the knee points for the ICP ($n=5$) and Doppler ($n=10$) estimators. 
Although OG does not rely on geometric correspondences, its measurements are typically less noisy than the Doppler velocity measurements, leading to a slower accumulation of drift between map updates. 
In contrast, while ICP provides the highest peak accuracy, its knee point is reached earlier because the computational cost of dense point cloud alignment scales poorly with frequency, unlike the OG estimator which has low computational overhead.
As a result, OG outperforms both ICP and Doppler at larger intervals, offering a more favourable trade-off and enabling a larger knee point.
Beyond the maximum intervals shown, accumulated odometric drift leads to localization failure; the lateral RMSE exceeds $0.2 \ \unit{\m}$, rendering the estimate unusable for real-world applications.

This comparative analysis highlights that at high update frequencies, the choice of odometry method has negligible impact on localization accuracy, as frequent map-based corrections mask the impact of drift.
Consequently, in scenarios where the computational budget allows for high-frequency localization, the use of a lightweight odometry method is preferred.
We demonstrate that the value of robust odometry is not in peak accuracy when localizing at every frame, but in its capacity to maintain stability when corrections are sparse.

\subsubsection{Trajectory Performance}
The structured \texttt{Suburbs} and \texttt{Industrial} routes generally favour all estimators.
In these environments, abundant static features ensure stable ICP correspondences while providing reliable inlier classification and radial velocity consistency for the Doppler estimator.
While the OG estimator performs well on the \texttt{Suburbs} route, it exhibits high longitudinal error on the \texttt{Industrial} route.
This degradation is likely due to the presence of snow since, unlike the other methods, OG is susceptible to wheel slip. 
In contrast, the sparser \texttt{Regional} route introduces a greater challenge containing a high-speed, open roadway with minimal roadside structure.
Consequently, all pipelines exhibit more rapid drift at larger localization intervals as the available information for corrections becomes increasingly sparse.

The \texttt{Tunnel} route highlights a failure mode for geometry-based methods, where the ICP estimator fails across all intervals due to longitudinal ambiguity. 
The lack of geometric constraints forces a heavy reliance on the motion prior to maintain localization.
Consequently, errors accumulate rapidly as the dead-reckoning interval increases, and the Doppler estimator fails to localize beyond $n=5$, once the accumulated odometry drift exceeds recoverable limits.
In contrast, the OG estimator maintains a knee point of $n=25$, even with increased error in the one sequence where the Doppler method fails entirely.
Despite an elevated mean error caused by that single difficult sequence, the overall data spread reveals a performance trend consistent with other routes. 
Notably, this trend demonstrates particularly low lateral and yaw error, as the gyroscope provides a stable orientation estimate that remains robust despite the absence of external geometric features.

Additional experiments were performed using the \texttt{Skyway} route \cite{Lisus2026BORRT}, though all pipelines failed to successfully localize. 
The environment provides insufficient geometric information to constrain a feature-based localization method. 
Consequently, the estimator becomes entirely reliant on the odometry prior, and without a near-perfect estimate to maintain alignment, the lack of distinct geometric features causes the scan-matching process to fail rapidly.

\subsubsection{Recommendations and Extensions}
This work demonstrates that our choice of odometry method can significantly reduce computational overhead in localization systems. 
The integration of the lightweight Doppler and \ac{OG} methods in a localization pipeline shows that fast estimations can be used when real-time performance is critical, leaving computational resources available for other tasks. 

Our results show that an odometry method does not need to be highly accurate to support highly accurate localization. 
While the Doppler estimator exhibits higher drift in odometry benchmarks \citep{YoonN4S}, it facilitates significant computational savings with negligible accuracy loss at high map-matching rates. 
Conversely, while the ICP estimator is highly accurate, it has a higher computational bottleneck, limiting real-time utility.
The OG estimator bridges this gap, providing a superior trade-off by enabling larger dead-reckoning intervals than Doppler without the overhead of ICP.

Ultimately, by controlling the localization frequency rather than localizing at every frame, efficiency can be optimized without altering fundamental algorithms.
This modular approach allows for customizing performance based on operational objectives, such as prioritizing the ICP pipeline for maximum precision or the OG and Doppler pipelines for resource-constrained environments.

Future work may explore dynamic localization intervals that adjust in real-time based on vehicle speed, route complexity, or estimator uncertainty. 
Such an adaptive framework would further optimize the balance between accuracy and efficiency across varying operational conditions.
\section{Conclusion}
\label{sec:conclusion}
This paper presents an evaluation of computational efficiency in lidar-based localization by leveraging diverse odometry methods within a unified pipeline. 
We demonstrate that integrating lightweight estimators, such as Doppler and OG, significantly reduces computational overhead while maintaining high localization accuracy when frequently map-matching. 
Our results show that the primary value of a robust odometry source lies in its ability to extend the dead-reckoning interval between map-matching updates, effectively balancing the trade-off between computational effort and accuracy.

For example, utilizing a performance-oriented ICP estimator and localizing every $5$ frames ($n=5$) yields a $27\%$ reduction in pipeline runtime compared to localizing every frame ($n=1$) while maintaining SOTA accuracy. 
The lightweight Doppler estimator allows for even greater efficiency; at $n=10$, it achieves an $80\%$ reduction in runtime with only a marginal sacrifice in accuracy compared to the ICP pipeline.
Finally, the OG estimator provides the most significant gains for resource-constrained applications, maintaining a stable estimate at $n=25$ and achieving a $91\%$ reduction in computational effort.
Furthermore, we evaluate these estimators in feature-sparse environments, such as tunnels, where geometric ambiguity causes ICP odometry to fail while OG maintains consistent performance.

Overall, this work reveals that localization frequency is a critical parameter for system optimization. 
By reducing reliance on localization corrections and relying on robust odometry for intermediate dead-reckoning, we can substantially improve the efficiency of existing localization pipelines, freeing critical onboard resources for other tasks within the AV stack.

\clearpage


\renewcommand*{\bibfont}{\footnotesize}
\printbibliography
\end{document}